# Learning-induced categorical perception in a neural network model

Christian Thériault[1], Fernanda Pérez-Gay[1,2], Dan Rivas[1], Stevan Harnad[1,3]

[1]*Université du Québec à Montréal*
[2]*McGill University*
[3]*University of Southampton, UK*

(address correspondence to: theriaultchristian@gmail.com)

**Abstract:** In human cognition, the expansion of perceived between-category distances and compression of within-category distances is known as *categorical perception (CP)*. There are several hypotheses about the causes of CP (e.g., language, learning, evolution) but no functional model. Whether CP is essential to categorisation or simply a by-product of it is not yet clear, but evidence is accumulating that CP can be induced by category learning. We provide a model for learning-induced CP as expansion and compression of distances in hidden-unit space in neural nets. Basic conditions from which the current model predicts CP are described, and clues as to how these conditions might generalize to more complex kinds of categorization begin to emerge.

## 1 Categorical Perception

*Categorical Perception* (CP) is defined by the expansion of the perceived differences among members of different categories and/or the compression of the perceived differences among members of the same category (Harnad 1987). In simple terms, if $a_1$ and $a_2$ belong to category A and $b_1$ belongs to category B, then $a_1$ is perceived subjectively as more different from $b_1$ than from $a_2$ even when the objective differences are equal. A clear consensus on the conditions generating CP has yet to be reached. According to the "Whorf Hypothesis" (Kay & Kempton 1984; Hussein 2012), the between-category separation and/or within-category compression that defines CP is cause by "language". (Things given different names look more different than things given the same name.) Without ruling out language as a factor in CP, genetics (i.e. biological evolution) and learning are also known influences. Biological factors are supported by the case of color perception: Two colors separated by the same distance in wavelength are perceived as more different when each color resides on separate sides of a color boundary. This CP effect is not a result of language, but a result of the brain's innate feature-detectors for colors (Jacobs 2013). Evidence (Bao 2015; Folstein et al. 2015; de Leeuw et al 2016; Goldstone et al. 2017; Edwards 2017; Cangelosi 2017; Pérez-Gay J, 2017) also suggests that CP can be induced by learning in other sensory domains. In such cases, perceived distances before learning are independent of category membership, but after the category has been learned, there is a perceived expansion of the distance between members of different categories and a compression of the distance between members of the same category.

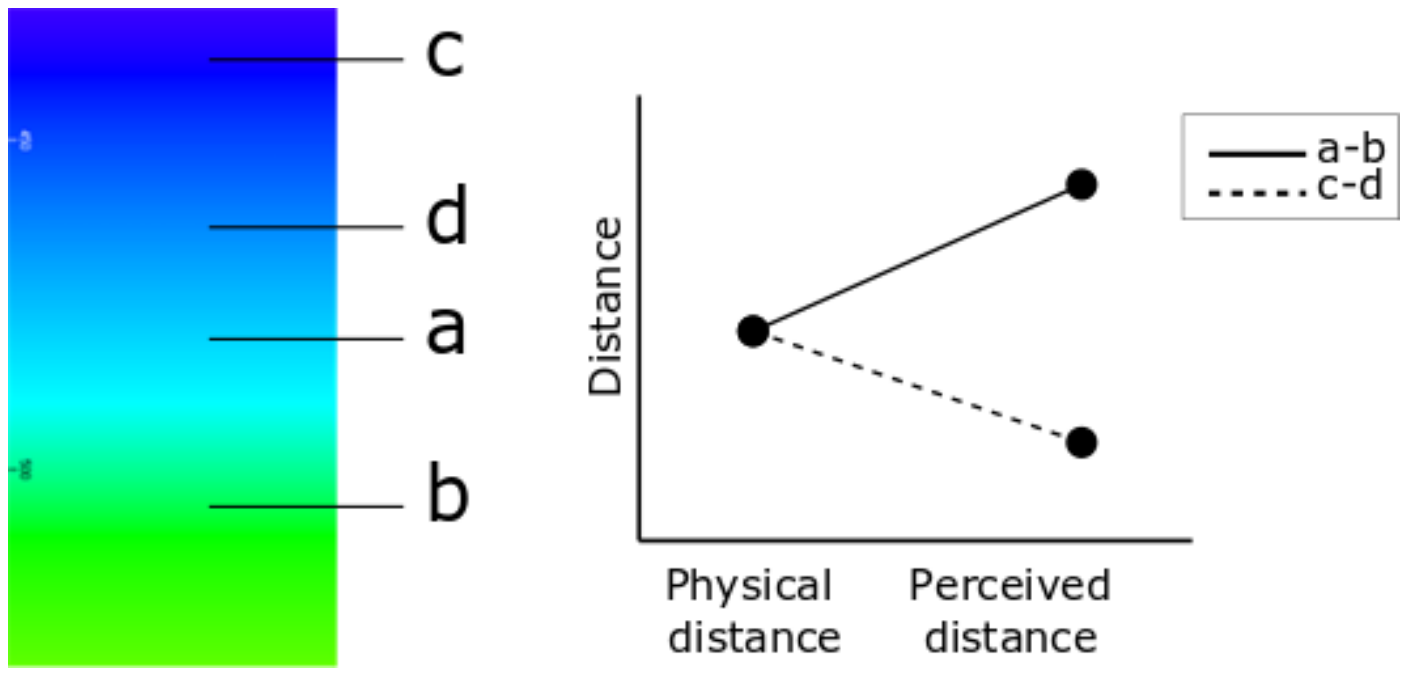

*Figure 1*. Hues crossing the boundary between two adjacent color categories in the spectrum are perceived as more distant than hues within the same category, even when physical (wavelength) distances are the same.

We propose a model for CP induced by category learning based on an idea originally proposed by Stephen J. Hanson (Harnad, Hanson & Lubin 1995), consisting of an unsupervised learning phase before category learning followed by a supervised category-learning phase. As a potential explanation for human experimental findings on learned CP our model examines the functional role of between-category separation and within-category compression in the internal representation of input stimuli before and after category learning. The explanation hinges on the number and proportion of category-invariant dimensions relative to the total number of dimensions in a simple binary stimulus feature-space.

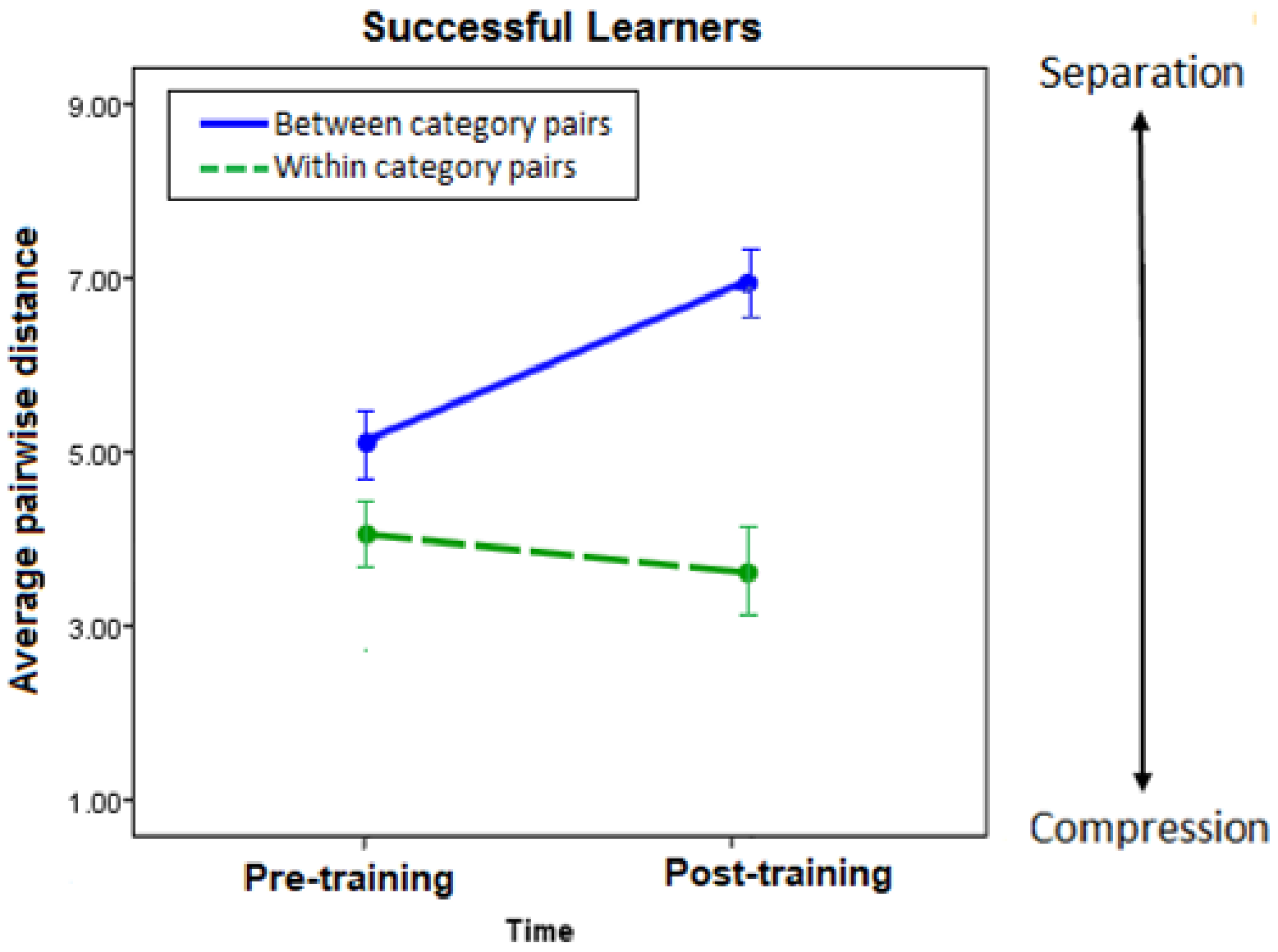

*Figure 2.* Human experimental data showing how pairwise stimulus dissimilarity judgments changed from before to after successfully learning the category through trial and error training with corrective feedback (adapted from Perez & al 2017).

### 1.1 Unsupervised and Supervised Learning

An initial analytical understanding of CP can be derived from two families of learning algorithms for pattern classification (Richard O & al, 2001, Bishop, 2006): unsupervised and supervised learning. Unsupervised learning extracts the *intrinsic structure* of data without any information about category membership, whereas supervised learning partitions data into categories on the basis of corrective feedback. A basic model of CP emerges when comparing data representations resulting from these two families of algorithms, with their different goals. By learning relevant features and discarding irrelevant ones, both families of algorithms can modify the distances among data-points. Unsupervised learning does so by converging on statistically relevant patterns inherent in the data-structure itself (feature frequencies and correlations), whereas supervised learning separates the data into categories based on external feedback on correct and incorrect categorization).

Within their respective families, multiple candidates (i.e., principal component analysis, independent component analysis, singular value decomposition, support vector machine, linear regression), share similar behaviors and could serve our purposes. Basic principles of category learning and CP can be expressed in particularly simple terms by principal component analysis (PCA) and linear discriminant analysis (LDA), both of which can be related to neural network models. PCA belongs to the family of unsupervised learning algorithms. Given a set of $n$ observations represented by $n$ column vectors $X = \{\mathbf{x}_1, \mathbf{x}_2, \mathbf{x}_3 \dots, \mathbf{x}_n\}$ in an $m$ dimensional space $\mathbb{R}^m$, PCA transforms these observations into a basis of $n$ uncorrelated principal components $B = \{u_1, u_2, u_3 \dots, u_n\}$ (Pearson, 1901). These components are found by minimizing the error of reconstruction when projecting data points on a subset of principal components. Figure 3 illustrates this principle for a two-dimensional space. The first component found by PCA captures the most variation in $X$. The following components successively capture the most variance in directions orthogonal to the previous components. As a result, PCA finds the lower dimensional linear space that best reconstructs the data with uncorrelated components. The components found by PCA can be defined by the eigenvectors of the covariance matrix $C$ with entries $(i, j)$

$$C_{i,j} = (\mathbf{x_i} - \bar{\mathbf{x}})^T (\mathbf{x}_j - \bar{\mathbf{x}})\,. \tag{1}$$

As only the maximum explained variance matters, PCA yields a good reconstruction but does not take category information into account. In fact, the projection on the main component may yield a representation that is worse than the initial representation in terms of category separation.

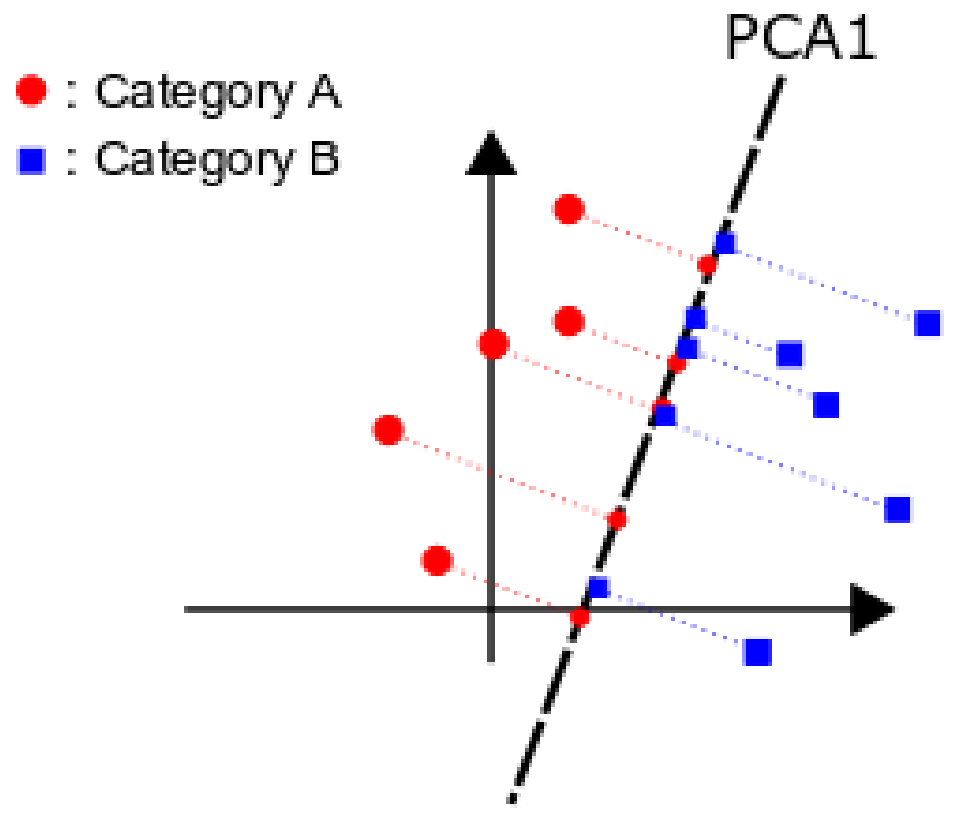

*Figure 3.* The projection line defined by the main PCA axis may yield an inappropriate categorisation space where categories are entangled.

Figure 4 illustrates this effect. Although the main direction found by PCA gives a good linear approximation of the data, the projective direction which best separates the categories is a different one. Linear discriminant analysis (LDA) provides this direction (Fisher, 1936), and belongs to the family of supervised learning algorithms.

Linear discriminant analysis (LDA) uses categorical information about the data to find a projecting vector (i.e. an hyperplane) which separates the categories by maximizing the ratio

$$S = \frac{S_b^2}{S_w^2} \ , \qquad (2)$$

where $S_b^2$ is the projected between-category-variance defined by the squared difference between projected category means and $S_w^2$ is the projected within-category-variance defined by the sum of the projected within-category-covariance. Figure 4 illustrates such a separating vector in relation to category space.

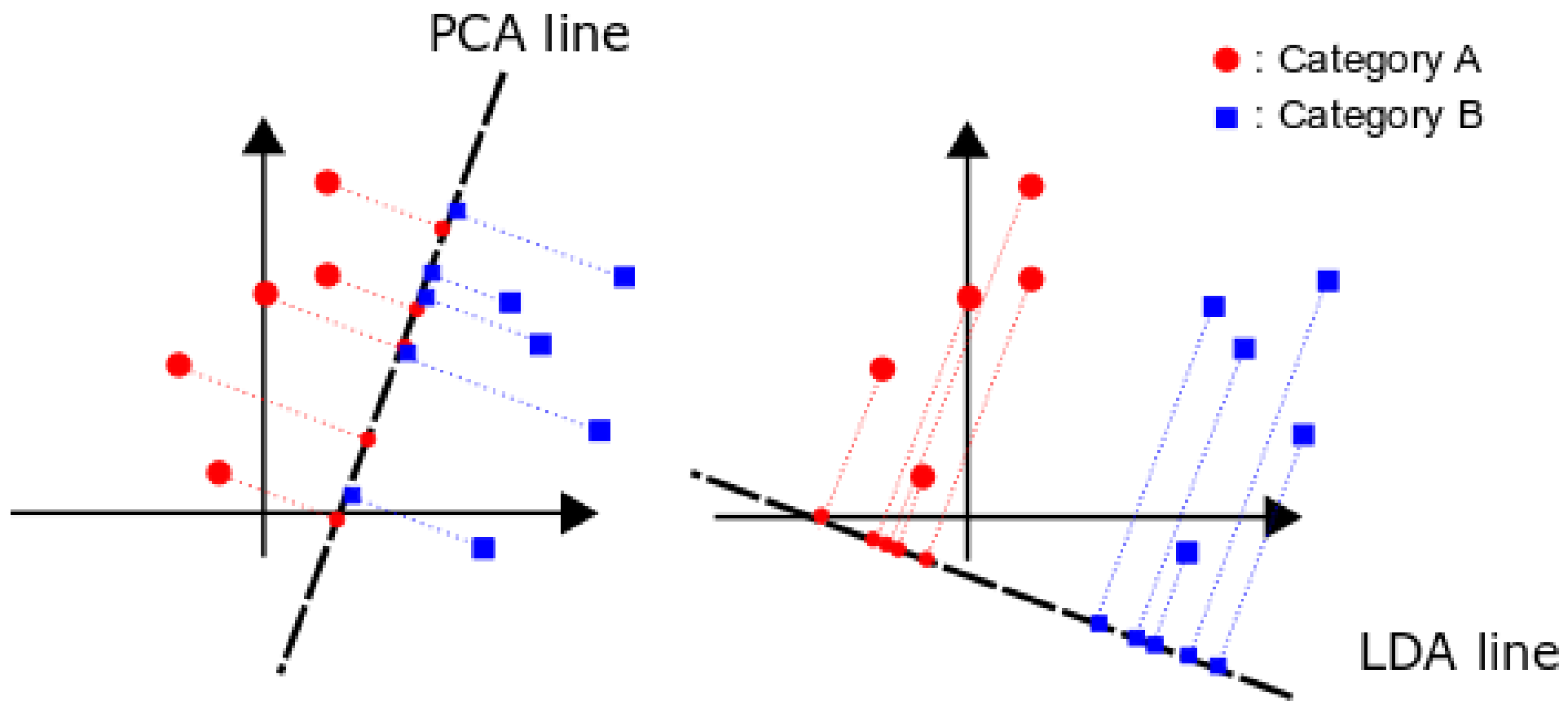

*Figure 4*. In this example, PCA is not a good projection line for categorization whereas LDA provides a clear separation between categories.

A comparison of unsupervised (PCA-like) and supervised (LDA-like) representation sets the stage for a simple first-approximation model of categorical perception. Referring to figure 4, there are no conditions preventing the entanglement of the category data points on the PCA line. But with the addition of category-membership information, projections on the LDA line will generate -- with respect to the initial PCA representation -- the separation/compression characteristic of CP. In such a case, as illustrated in figure 5, we observe the expansion of the distances between members of categories A and B, and the compression of the distances among members of the same category (A or B).

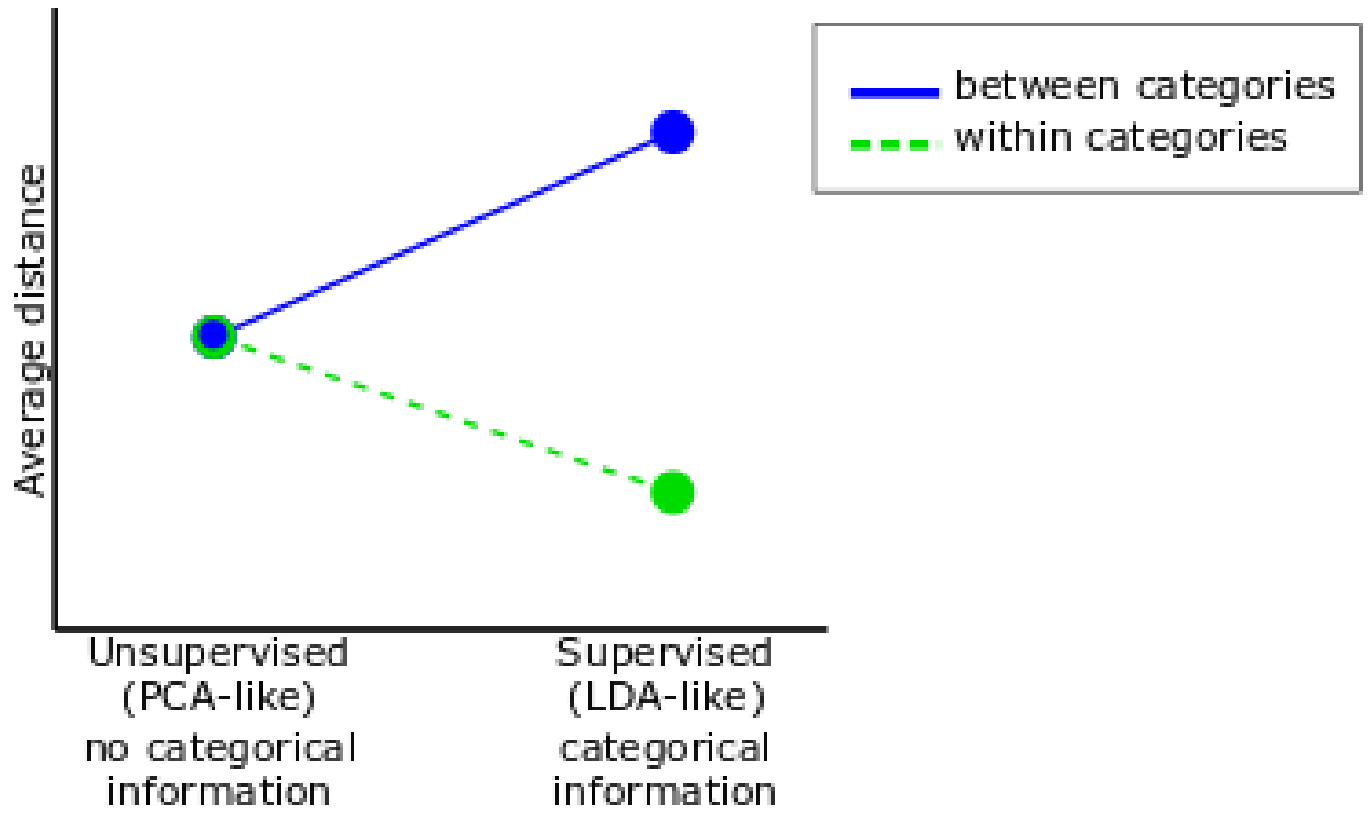

*Figure 5* Example of between-category distance and witin-category distance when measured as linear projection on unsupervised and supervised component (as in Figure 2).

What is of particular interest for modeling CP, is that PCA and LDA (and equivalent candidates in their respective families) can be related to the activation layers in a neural network during two phases of learning: an unsupervised phase (PCA-like) and a supervised phase (LDA-like). A representation equivalent to PCA is generated by an autoencoder with a single hidden layer (Boulard,H 1988 ; Baldi,p 1988). Also known as auto-associative memories (Hopfield, 1982), these networks learn to represent and regenerate inputs irrespective of category membership. By learning the relevant subspace (i.e. relevant features), autoencoders develop internal representations which enable reconstruction of incomplete inputs. In terms of the performance of human subjects, this corresponds to becoming familiar with stimuli through repeated exposure, with no particular contingencies or instruction regarding categories or categorization.

A representation related in spirit to LDA is obtained by changing the unsupervised learning rule of the net to a supervised one by introducing corrective feedback about category membership. Supervised neural networks use error-corrective feedback (LeCun, Bengio, & Hinton, 2015) to learn connection weights which define a subspace separating the categories. The activation layer of such networks can be described as a *homeomorphism*, stretching and thinning the stimulus representation manifold, with the goal of separating the categories. For human subjects, this would correspond to subjects interacting with an object and receiving feedback (i.e., or reinforcement or "supervision") on correct categorization until few or no errors are made. The next section describes a basic neural network based on these principles.

## 2 Neural Network Model for Categorical Perception

This section describes a neural network that generates a CP effect as a result of (1) mere exposure to the stimuli with no categorical information followed by (2) category learning with error-corrective feedback. The network is described as a single architecture in which the connection weights are first learned in an unsupervised manner, followed by a supervised phase in which categorical information becomes available to the network.

### 2.1 Auto-Encoding Phase

A denoising auto-encoder (Vincent, 2008) network learns to generate as output the same training stimulus it receives as input (it learns to map inputs onto themselves). Through exposure, the network learns a representation from which it is able to regenerate learned examples **x** when presented with partial, incomplete or noisy examples $\hat{\mathbf{x}}$. The forward and feedback output of layer **h** and layer **x** are respectively given by

$$\mathbf{h} = f(\mathbf{a_h}) \ , \qquad \mathbf{a_h} = \mathbf{Wx} + \mathbf{b_h} \tag{3}$$

and

$$\mathbf{x} = f(\mathbf{a_x}) \ , \qquad \mathbf{a_x} = \mathbf{W}^T\mathbf{h} + \mathbf{b_x}$$

where ***f*** is a nonlinear output function, **W** is the connection weights between the layers, and **b** is an activation bias.

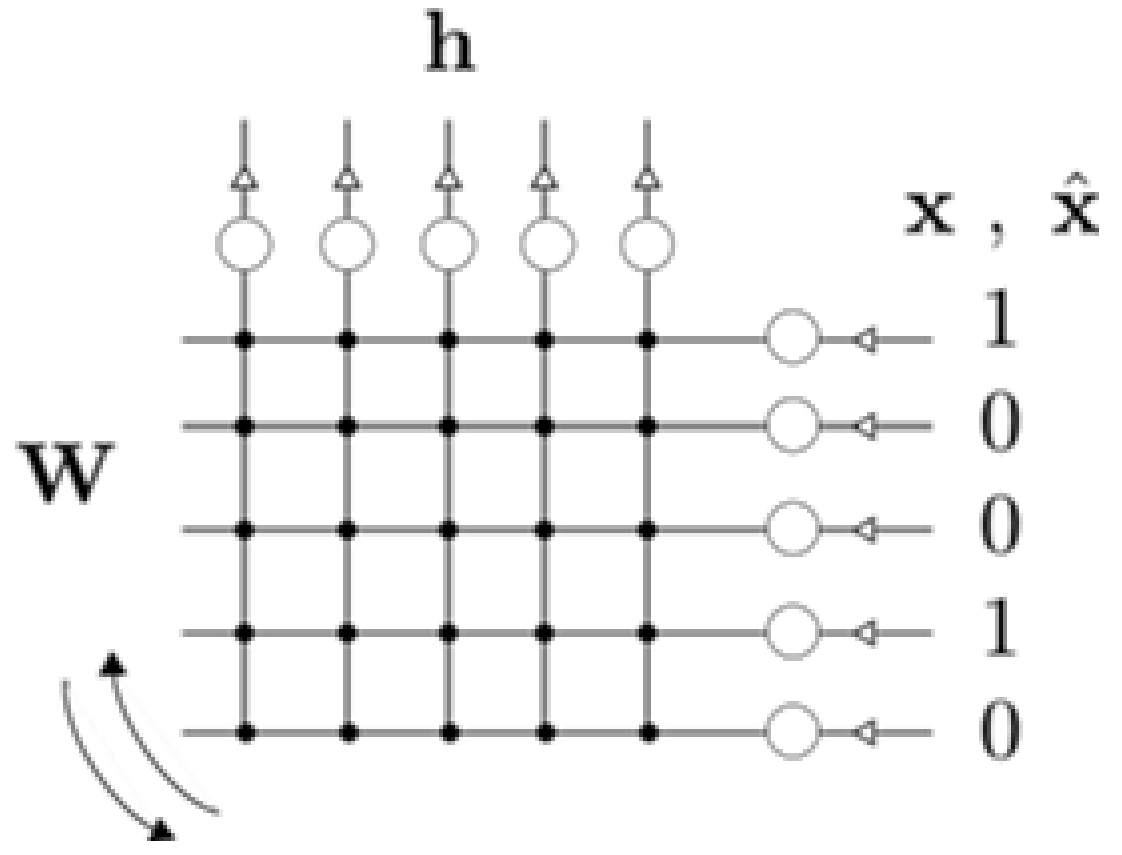

*Figure 6* Basic autoencoder architecture with a symmetrical weight matrix.

The connection weights are trained by repeated presentation of noisy examples $\hat{\mathbf{x}}$. During the learning process, **W** is modified by error gradient backpropagation to minimize the error $E = ||\mathbf{x} - f(\mathbf{h}(\hat{\mathbf{x}})||^2$ obtained when the network is attempting to regenerate examples **x** from noisy examples $\hat{\mathbf{x}}$. A purely unsupervised auto-encoding can be achieved by using less network units than input dimensions, forcing the network to learn relevant features even in the absence of noise during training.

### 2.2 Supervised Phase : Classification Network

Once the auto-encoder has learned a proper space of representation, this is passed forward to a second level of representation where the categories are learned using error-corrective feedback. Connection weights $\mathbf{W_1}$ , $\mathbf{W_2}$ are modified through gradient backpropagation based on the net's error -- the difference between its output and the correct category name -- strengthening correct connections and weakening incorrect ones.

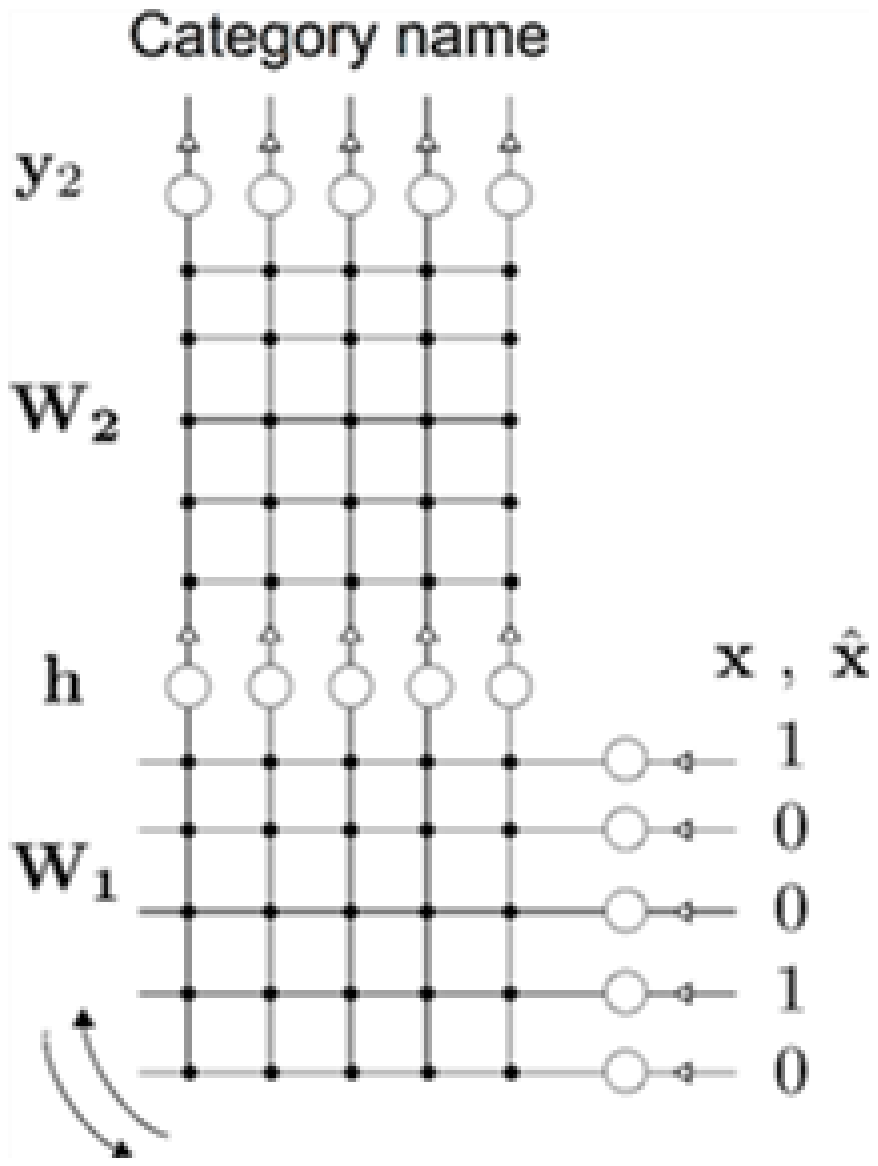

*Figure 7*. Basic classification network with one hidden layer initialized with autoencoding.

Stimulus representations are measured on the internal layer activation $\mathbf{a_h}$ first after the auto-associative phase and then after the supervised learning phase. The separation-compression effect is evaluated by measuring distances between these representations.

### 2.3 Separation-Compression Measure

To assess the separation-compression effect of category learning, let $h^u$ and $h^s$ be respectively the unsupervised and supervised representation on the inner layer of the network. One can define the average Euclidean distances $D_A$ , $D_B$ *within* each category and the average distance $D_{A,B}$ *between* categories, in the case of supervised and unsupervised representations, with

$$D_A^u = 1/T_a \sum_{i \in A} \sum_{j \in A, j>i} \|h_i^u - h_j^u\| \quad , \quad D_B^u = 1/T_b \sum_{i \in B} \sum_{j \in B, j>i} \|h_i^u - h_j^u\|$$

$$D_A^s = 1/T_a \sum_{i \in A} \sum_{j \in A, j>i} \|h_i^s - h_j^s\| \quad , \quad D_B^s = 1/T_b \sum_{i \in B} \sum_{j \in B, j>i} \|h_i^s - h_j^s\| \tag{4}$$

$$D_{A,B}^u = 1/T_{ab} \sum_{i \in A} \sum_{j \in B} \|h_i^s - h_j^s\| \quad , \quad D_{A,B}^s = 1/T_{ab} \sum_{i \in A} \sum_{j \in B} \|h_i^s - h_j^s\|$$

where, $T_a, T_b, T_{ab}$ are the corresponding total number of pairwise distances. The between-category $SP_b$ category separation and the within-category $SP_w$ compression can be measured as

$$SP_w = \frac{(D_A^s + D_B^s) - (D_A^u + D_B^u)}{2} \qquad (5)$$
$$SP_b = D_{A,B}^s - D_{A,B}^u$$

## 2.4 Simulation 1

There is no a-priori way to equate the particular parameters of the simulations reported in this paper directly with those of any human experiment. The neural model described above is not a model of visual or auditory perception (which would require at least a convolutional neural network); it is a general fully connected model (every unit is connected to every other unit) where the dimensions processed by the network are meant to be purely formal and arbitrary. Nevertheless, for a first attempt at modeling and understanding the kinds of findings observed in humans, the experiment reported by *Perez-Gay J & al* (2017) provide us with a set of formal stimulus definitions with no obvious perceptual structure. These stimuli are explicitly defined in terms of the number $k$ of category-invariant features relative to the total number of features $N$.

## 2.5 Stimuli

The stimuli for this initial experiment are defined similarly to those of *Perez-Gay J & al.* (2017) who reported learning-induced CP (between-category separation, and sometimes also within-category compression) using randomly generated stimuli composed of small binary image patches. These stimuli were composed of 6 local micro-features consisting of binary image pixel grids, $k$ of them co-varying with membership in each category. The other $6 - k$ dimensions were random, having no relation to category membership.

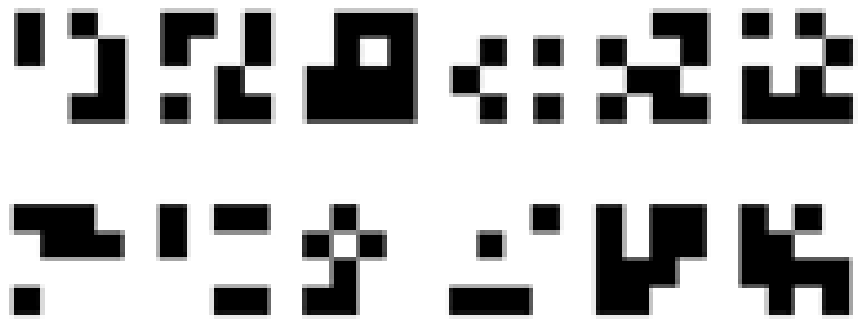

*Figure 8.* Binary visual micro-features used to create the stimuli in Perez-Gay J & al *(2017*). In the present formal simulation, the stimuli are not arrays but vectors with N binary components.

The stimuli in the two categories, A and B, were larger grids, consisting of $M \times M$ dimensional binary images composed out of the $N$ micro-features randomly distributed in equal proportions. For this experiment, a very approximate simulation of the stimuli used in *Perez-Gay J & al.* (2017) on human subjects was generated by computer as shown in Figure 9, using 4 micro features.

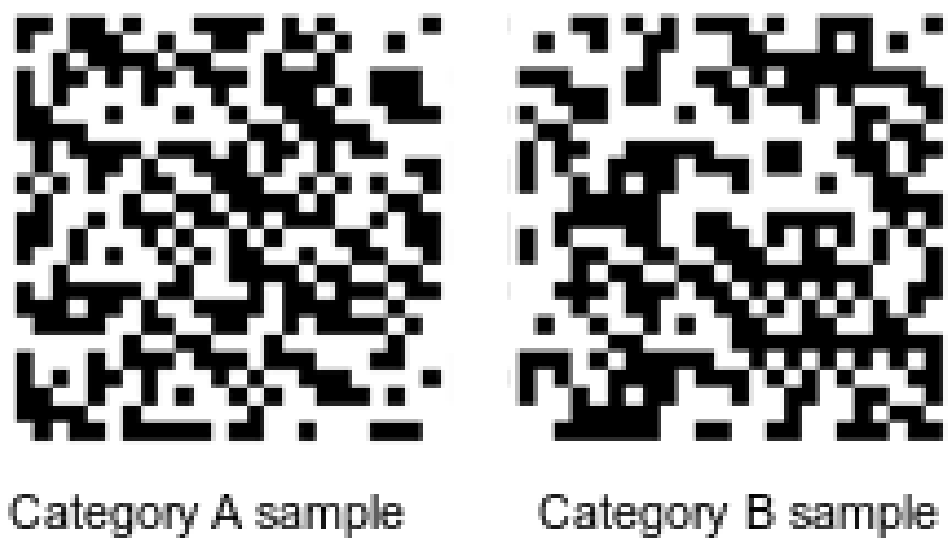

*Figure 9* Sample images from category A and B.

### 2.6 Results

Classification rates of 99 % or more are achieved when the nets are given enough training trials. But what is of interest is the correlates of having successfully met the categorization criteria. As a simulation of the human experimental results shown in Figure 2, Figure 10 shows the effect of the number $k$ of covariant features on the separation-compression measures $SP_w$ and $SP_b$, for $N = 4$ micro features, and decreasing values of $k$. An increase in between-category separation and within-category compression is observed as $k$ decreases.

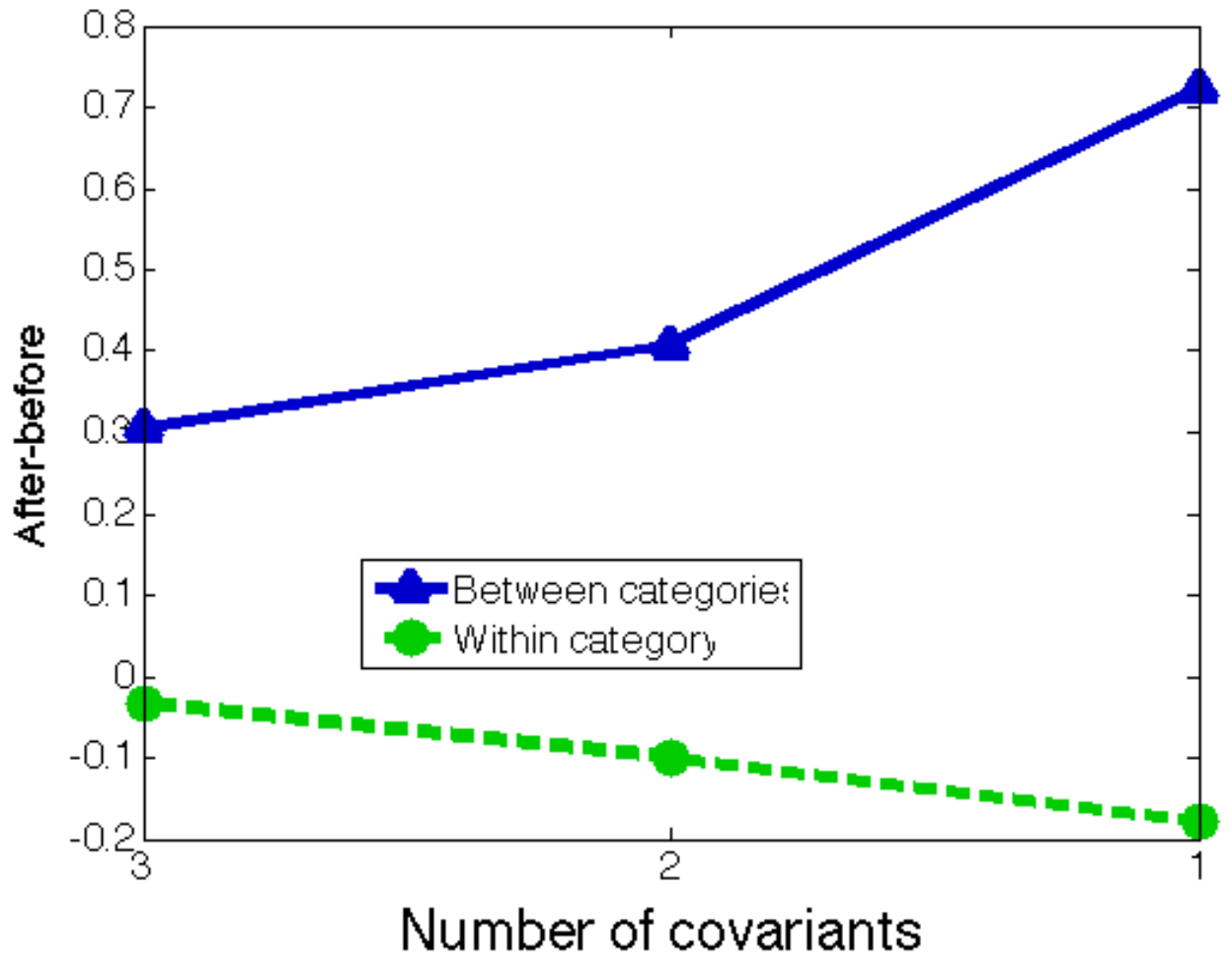

*Figure 10*. Evolution of between-category separation and within-category compression with training. The vertical axis gives the separation-compression measures $SP_b$ and $SP_w$, as defined by equation 3 (difference between average pairwise distances measured following unsupervised learning and then following supervised learning).

The effect of the relative number of category-covariant features ($k/4$) on separation-compression already provides clues as to what might underlie the experimental results on humans. The network is able to capture statistically relevant patterns because of the redundancy created by the $k$ covariant features. We observe an increase in between-category separation and an increase in within-category compression as $k$ increases. The $N$ micro-features (figure 8) used to construct the stimuli (figure 9), however, are spatially and randomly distributed. This prevents making an explicit one-to-one mapping between the $k$ covariant features with the network dimensions because units in the network are not invariant under transformations in spatial position [1]. In other words, the spatial randomization of the micro-features is not well-suited to identifying the influence of category-covariant features in a network that is not spatially invariant. To better explore the impact of the number of category-covariant features in a purely abstract manner, independent of spatial topology (i.e. no requirement of spatial invariance), the next simulation defines a stimulus space where each dimension directly represents an arbitrary binary feature**.**

[1] For human experiments in *Perez-Gay J & al.* (2017), equating features for salience and detectability is also a problem.

## 3 Simulation 2

### 3.1 Stimuli: Category Definition

The stimuli used for this simulation are $N$ dimensional binary/bipolar vectors $\mathbf{x} \in \{-1,1\}^N$. Each constituent dimension is interpreted as a binary feature. The stimulus space thus corresponds to the corners of an $N$ dimensional hypercube. Two categories are defined by randomly selecting a subset of $k$ dimensions that co-vary with category membership. Each of these $k$ dimensions is randomly assigned -1 or 1 for one category and its binary complement (1 or -1) for the other category. These $k$ dimensions are thus the (conjunctive) covariants defining the categories. The other $N-k$ dimensions are completely random, having no correlation with category membership.

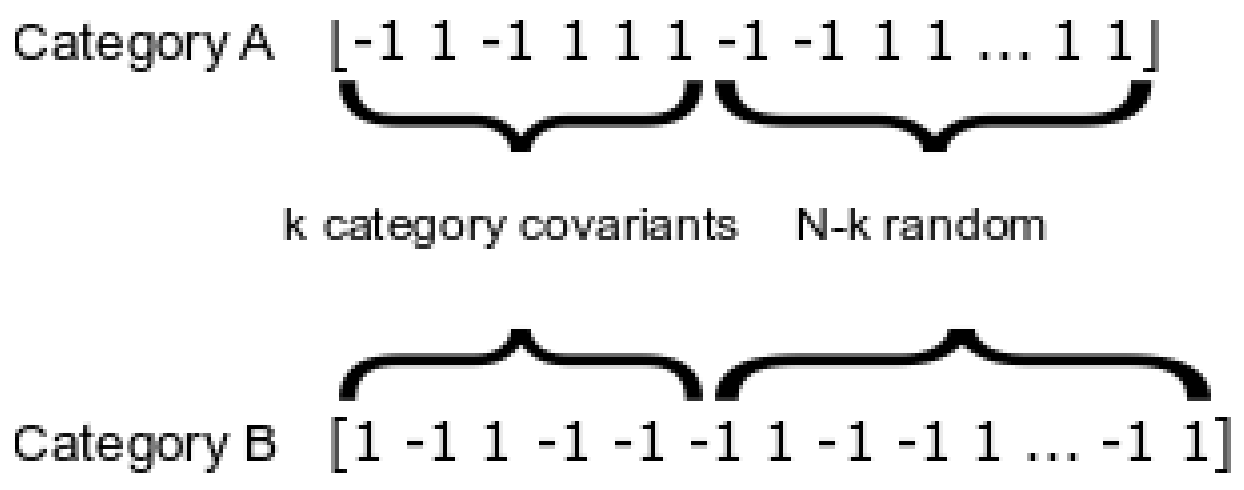

*Figure 11*. Binary (bipolar) stimuli defined by a conjunctive rule on the constituent dimensions.

### 3.2 Design of Simulations

Each simulation consists in training and testing the network with 1500 examples of binary vectors as defined above, for two categories, $A$ and $B$. A total of 5 simulations was done, each time with new randomized values defining the categories and the initial network connection weight matrices $\mathbf{W}_1$, $\mathbf{W}_2$. For each simulation, the auto-encoder and the classification layers were trained until a 99% success rate was reached. The representation $h$ of each stimulus is first recorded on the inner layer of the network following the unsupervised training phase; then following the supervised training phase the measures $SP_w$ and $SP_b$ are computed.

### 3.3 Results

The first criterion that should be satisfied, irrespective of any separation of compression, is the ability of the network to categorize stimulus space. Averages of 99 % classification success or better are achieved when given sufficient examples. But what is of interest is the correlate of having reached the categorization criterion. Figure 12 shows the evolution of between-category separation and within-category compression for values of $N = 10, 20, 50, 100$ and corresponding decreasing values of $k$.

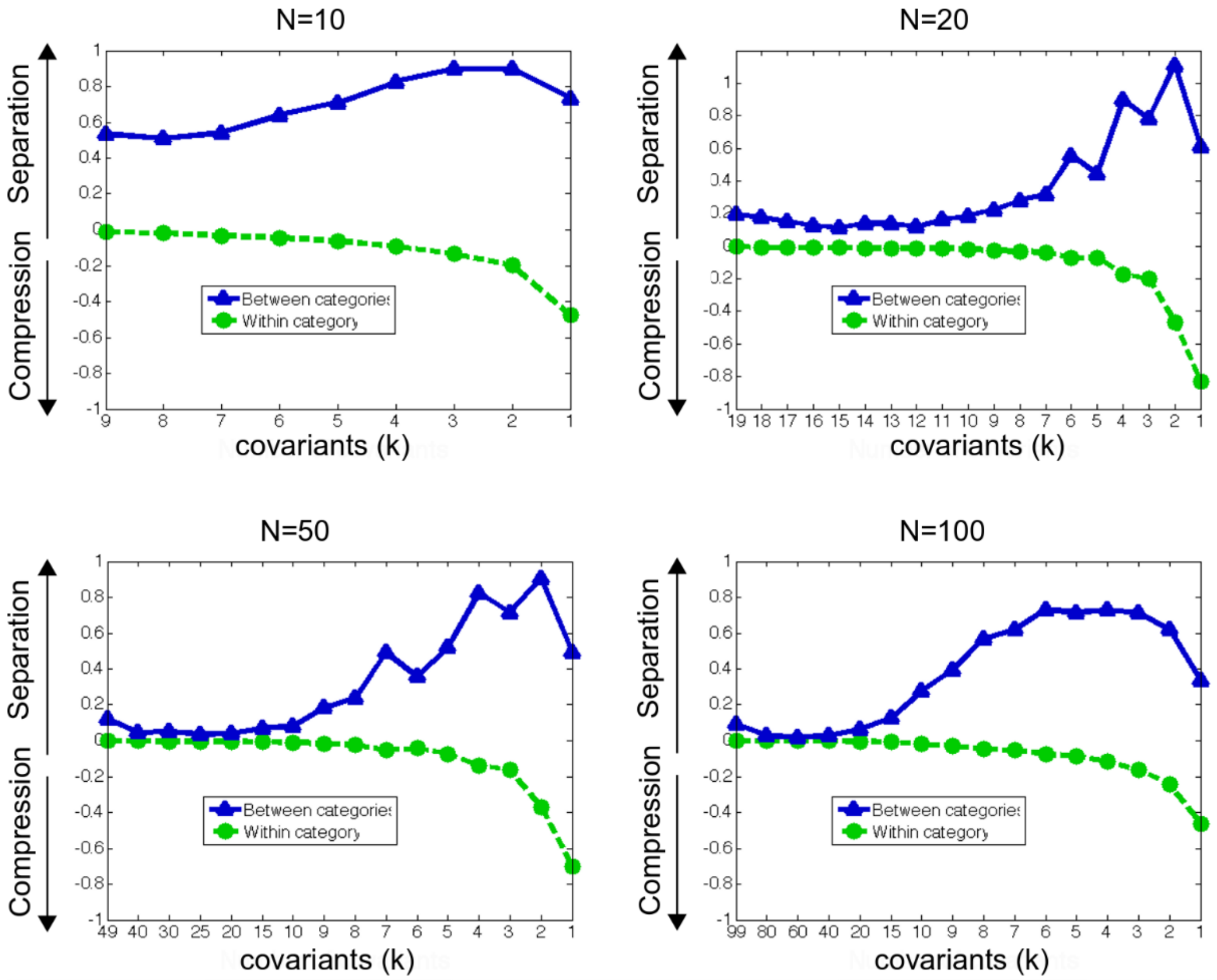

*Figure 12*. Evolution of between-category separation $SP_b$ and within-category compression $SP_w$ for different numbers of dimensions ($N$) and covariants ($k$). As defined by equation 5, the vertical axis gives the difference between distances computed after supervised learning and after unsupervised learning.

As the number $k$ of category-covariant features decreases, a general increase of between-category separation and within-category compression is observed. When learning the categories the network allocates more weight to category-relevant (covariant) dimensions and less weight to category irrelevant (i.e., noisy, uncorrelated) dimensions. This is what creates the observed increase in between-category separation and within-category compression. *Why?* With higher values of $k$ (large proportion of conjunctive covariants), the categories are already far apart (i.e. very different) structurally. Large distances between categories create a high variation in the data along category-separating directions. Because non-supervised representations favor directions of high variance, this makes the non-supervised representation a near optimal solution for categorization when the proportion of covariants is high: there is little or no difference between the unsupervised and the supervised values between and within categories for high values of $k$.

But as $k$ decreases, the unsupervised learning becomes less optimal for categorization, and separation is then generated by supervised category learning.

W*hy do we observe a drop in between-category separation correlated with an increase in within-category compression at the extreme low values of* $k$*?* As the value of $k$ decreases, more irrelevant (i.e. noisy) dimensions are eliminated or weakened by category learning. This progressive dimension reduction with decreasing $k$ creates an exponential decrease in the volume of space occupied by the categories after categorization and hence a commensurate decrease in the average pairwise distances: everything is measured inside a smaller space. This explains the more pronounced within-category compression at the very lowest values of $k$, correlated with weakened (but still present) category separation. This intensified compression and weakened separation near the lowest values of $k$ is simply the effect of dimension-reduction on Euclidean distances. This is a natural consequence of the model used in this simulation. Ascertaining whether this property is relevant for explaining human results, however, will still call for a lot of systematic testing and comparison.

If instead of looking at the compression and separation measures of equation 5 one looks directly at the average pairwise distance using equation 4, a clearer pictures emerges. Figure 12 shows the average pairwise distances measured on the network's inner layer after unsupervised learning ($D_A^u + D_B^u, D_{A,B}^u$) and after supervised learning ($D_A^s + D_B^s, D_{A,B}^s$) for $N = 50$ and decreasing values of $k$.

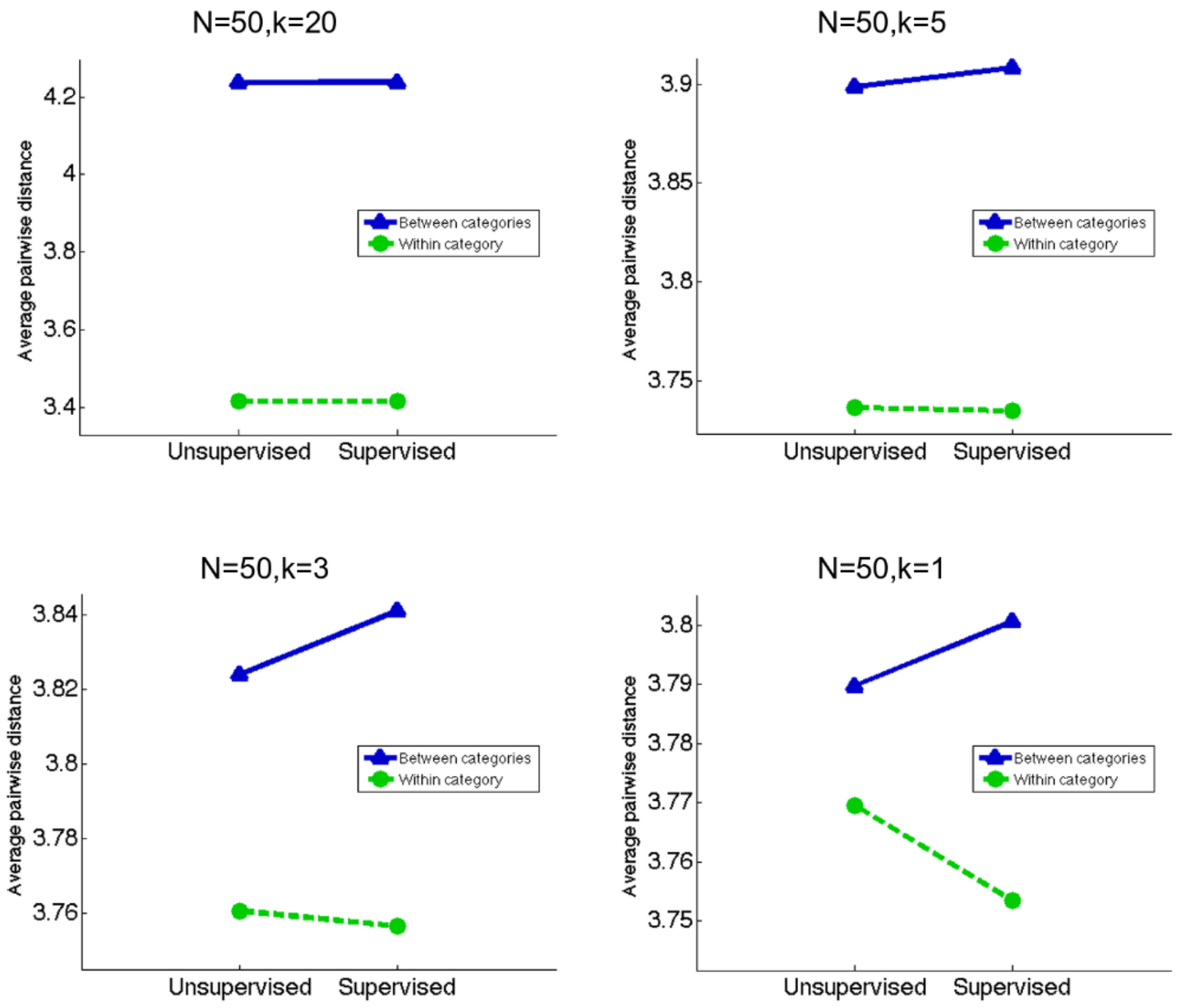

*Figure 13*. Average between-category pairwise distance and within-category pairwise distance measured first after unsupervised learning and then after supervised learning. Vertical axis in on logarithmic scale.

The range of values on all distance axes indicates what is expected from the stimulus definition and from the model. In *stimulus space* (i.e. physical space), for high values of k, the stimulus definition is such that categories are far apart and compressed (most dimensions are identical). But for low values of $k$, categories are closer and larger in volume. This results in a larger initial difference between pairwise within-category distances and between-category distances at high values of $k$. But Figure 12 also shows that as $k$ decreases, the difference between the initial within-category distances and the between-category distances following the unsupervised phase (before the supervised phase) decreases. One would expect that at $k = 1$, the minimum number of covariants needed to create two categories, the initial difference would be zero. Figure 12 shows a decrease in initial difference, but it does not quite reach the expected zero at $k = 1$. *Why?* The stimulus definition (Figure 11) creates a linear space between categories. This forces the average pairwise distance between categories to be larger than the average pairwise distance within categories. But one can easily create a zero initial difference in the average pairwise distance at $k = 1$ by *narrowing* the linear space between categories, making categories closer to each other. Figure 13, illustrates a stimulus definition that will generate a zero initial difference. This definition is the same as the previous one, except that the categories are brought closer to each other at $k = 1$. It is equivalent to reducing the salience of the only covariant feature at the hardest level of categorization. It follows that just as in Pérez-Gay et al's (2017) human experiments, the *a-priori salience of the perceptual features* is a parameter that needs to be closely examined and controlled as much as possible.

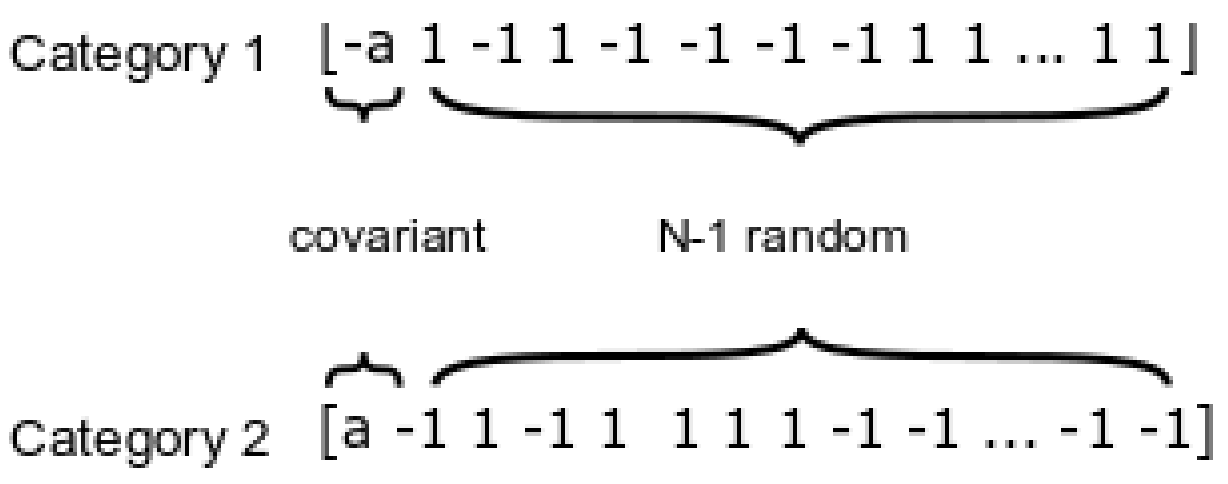

*Figure 14*. For $a < 1$ the linear space between categories at $k = 1$ is narrowed, reducing the average pairwise distance between categories. This can be thought of as lowering the salience of the only covariant feature (i.e., the most difficult level).

Figure 15 shows the average pairwise distance measured on the network inner layer after unsupervised learning and after supervised learning, for values $a = 0.6$ , b = 1 and $N = 20$.

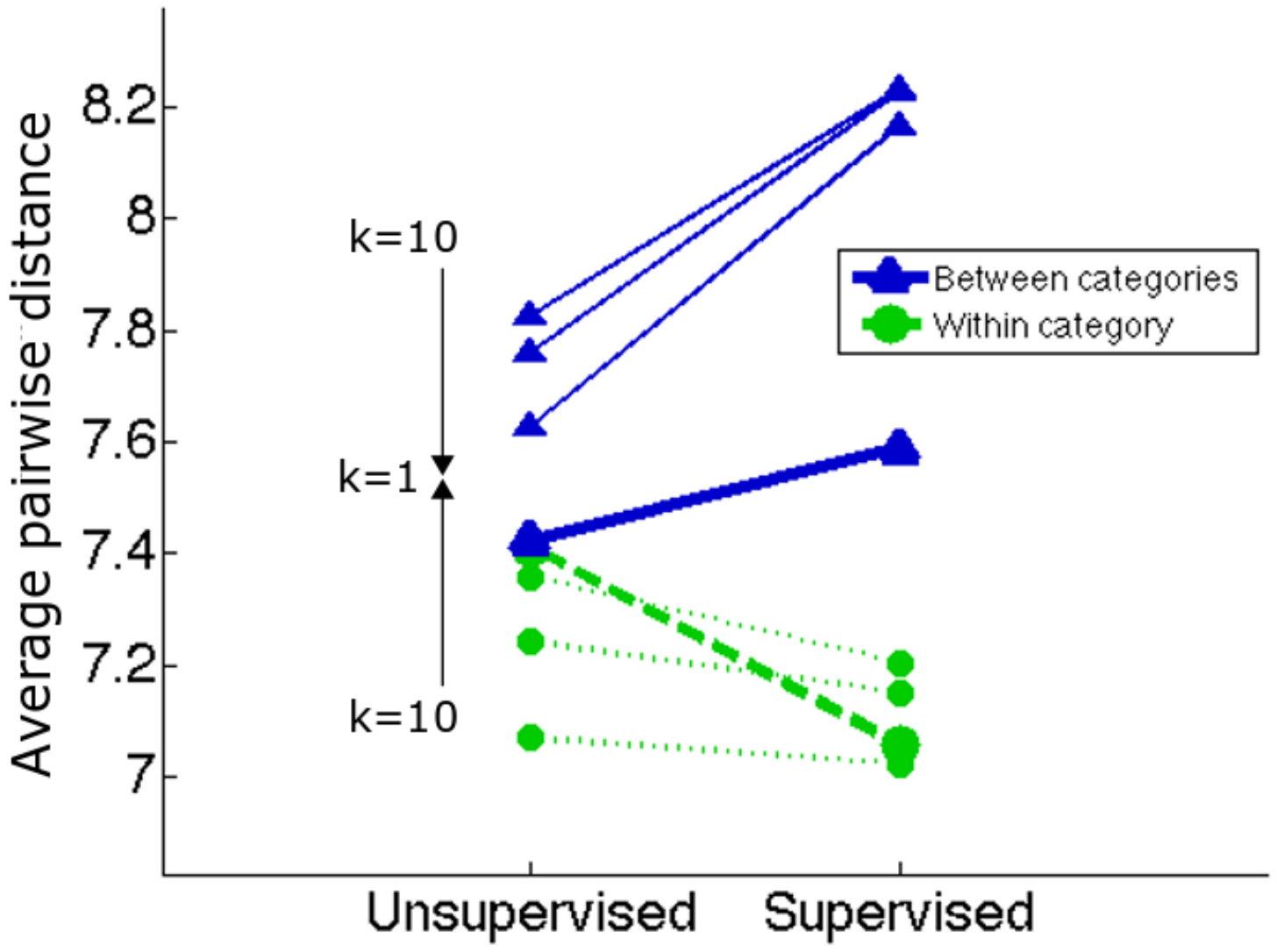

*Figure 15*. As the number $k$ of covariants decreases, there is less difference between the initial average pairwise distance between and within categories, with zero initial distance at $k = 1$. Results are shown for $N = 20$. Vertical axis in on logarithmic scale.

What the model predicts is again increased between-category separation and increased within-category compression with decreasing values of $k$, except at the lowest values of $k$ (i.e. near $k = 1$). This time, however, when $k = 1$, the initial difference is zero, indicating that average between- and within-category distances are equal when the number of category covariants is minimal.

## 4 Analysis and Discussion

In the model introduced above, two factors determine whether CP separation-compression occurs: initial encoding and dimensionality.

### 4.1 Initial Encoding

If mere exposure to stimuli, without categorical information (i.e. error correction, feedback), provides sufficient separation to achieve categorisation by simply associating a label with each category after the unsupervised learning, the above model predicts that no separation-compression (CP) will be observed. Consider the following analogy. Someone has never seen cats or cars and is presented with pictures of them. After repeated unlabelled exposure to hundreds of different examples of each, the viewer is told the category name associated with each picture, namely, "cat" or "car". Cats and cars are already so different that being told their name does not improve the ability to detect which is which. No supervised learning is necessary; just a few trials to learn what to call each. According to our model this means that in cases where the unsupervised separation is already along categorical lines, there is less separation-compression in going from the unsupervised to the supervised representation. In such cases, further training might be unnecessary, depending on factors such as weight-initialization at the top layer (i.e. random initialization) and how high the performance criterion (percent correct) for successful categorization is.

Figure 16 illustrates this principle. On the right, the subspace that is learned without supervision and best describes the data (i.e. PCA) is also a category-separating subspace. In such a case, the addition of categorical information (i.e. LDA) does not change the learned subspace, and little or no separation-compression will be observed when going from one representation to the other.

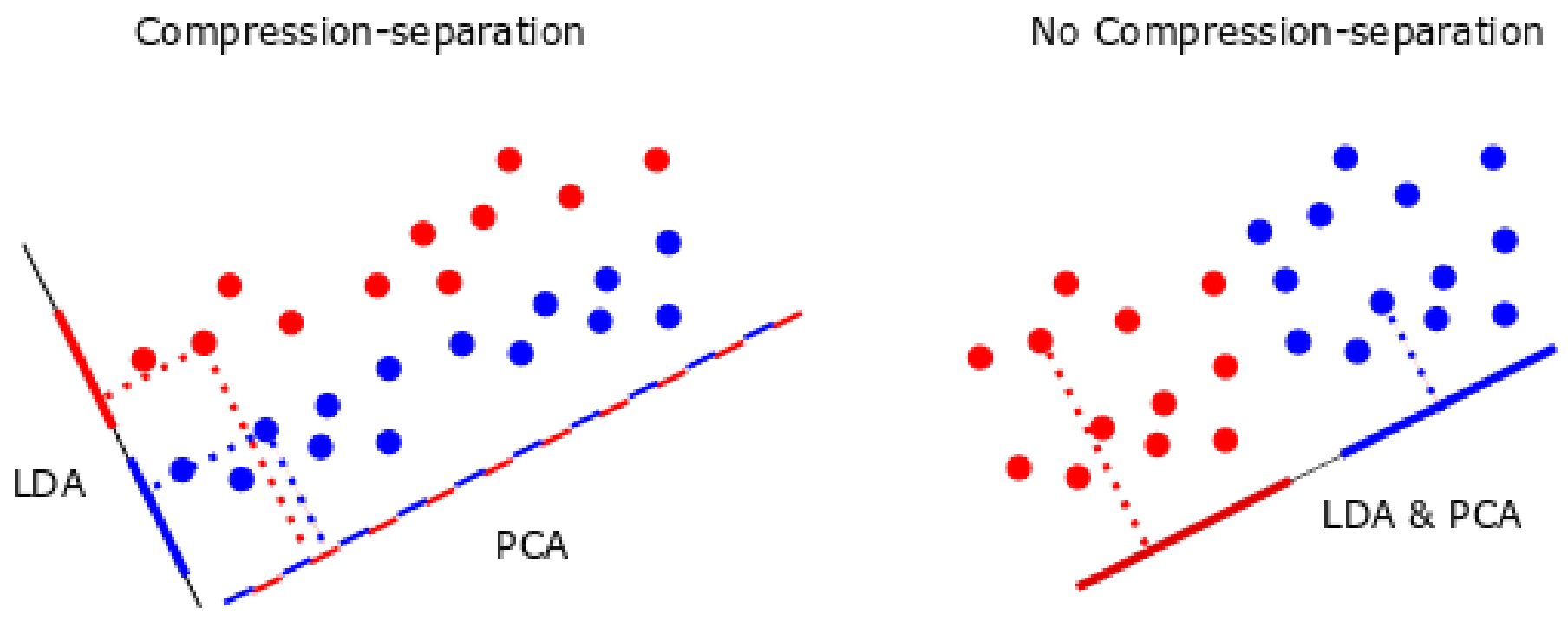

*Figure 16* If the unsupervised separation already separates the categories, little or no further separation/compression is needed.

This pattern is observed in the results of Simulation 2. With large values of $k$ (large number of conjunctive invariants), the distance between the categories is large and the network needs almost no modification of its unsupervised state to achieve error-free categorisation. There is still some separation (see figure 12), but less than for lower values of $k$. Large values of $k$ generate almost no change from unsupervised to supervised learning in terms of between-category distances. The analogy with human learning is that categories that differ in enough features need less supervision to learn their names: hence the perceived differences between and within categories do not change much as a result of supervised learning (Pérez-Gay J, 2017).

### 4.2 Dimensionality

The total dimensionality of the space and the proportion of category-relevant dimensions plays a crucial role when we compare unsupervised and supervised internal representations. Starting from the full set of dimensions of variation in stimulus space, unsupervised learning first reduces total dimensionality by bringing into focus the intrinsic structure of the data space based on relative feature frequencies and feature covariation. Then the supervised learning partitions this intrinsic structure into categories based on corrective feedback as to what belongs in each category. Dimensions that separate the categories get more weight, causing between-category separation, and dimensions that are irrelevant to categorization get less weight or are suppressed altogether, causing within-category compression. But as a side-effect, the dimension reduction caused by category learning creates an exponential decrease in the volume occupied by the categories as $k$ decreases. Smaller volume implies reduced Euclidean distances. This side-effect results in even stronger compression and a weaker separation at the extreme lowest values of $k$.

The dimensionality of the feature space defining categories also has a direct effect on the categorisation network's performance, a problem known as the curse of dimensionality (Hughes, 1968, Trunk, 1979). For a fixed amount of training data, as the dimensionality of the stimulus-space grows, the sparsity of the space increases exponentially (i.e., vaster space, fewer data points). This increases the probability that any direction is a category-defining direction. Consequently, for higher values of $N$ and a fixed number of training data points, the non-supervised direction (i.e., the autoencoder network) has a higher probability of being a category-separating direction. High values of $N$ and $k$ also mean high variation in category defining directions. Both these factors make the unsupervised representation a good candidate for category separation.

These principles may generalize to more complex category definitions as discussed in the concluding remarks. Indeed, deeper neural architectures build successive representations where initially tangled representations progressively become untangled, allowing linear separation of categories in the final layers of representation, thus potentially providing the conditions for inducing a CP effect for any category structure, regardless of its complexity.

## 5 Conclusion

Apart from modeling successful category learning, our interest is in finding a functional model for learning-induced CP. This first pass is based on characteristics of unsupervised and supervised learning. The goal is not to find a model in which learning always induces CP; this is not observed in human learning. The goal is to predict under what conditions learning will induce CP and to explain how, and why. The model identifies some of the conditions under which CP does and does not occur. The binary rule of Simulation 2 defines N-component stimulus vectors in which each of the N dimensions represents a binary feature. The stimuli and categories used in the simulations are all simple dichotomies (only two categories) and category membership is based on *conjunctive features:* all the members of one category have one of the values (0 or 1) on all k category-covariant binary features or dimensions and all the members of the other category have the complementary binary value (1 or 0) on all k. A stimulus is a member of a category if and only if all k category-covariant features are present. A fully connected network (i.e. a network in which every unit is connected to every other unit), as used in these simulations, can learn the categories by increasing the weights on category-covariant dimensions and reducing the weights on category-irrelevant dimensions. In visual perception, invariant features may or may not have a fixed spatial locus. In Simulation 1, an approximation to experiment 3 of Perez-Gay et al (2017), the stimuli are defined by a conjunctive rule based on visual micro-features (local) that are spatially and randomly distributed rather than having a fixed spatial locus. In learning the conjunctive rule of Simulation 1, the spatial invariance properties of convolutional networks (a network composed of local filter providing invariance to position) would perform better than the fully connected networks described in this paper. Future simulations will go on to test convolutional models on more realistic and more complex categories, including non-binary features, multiple rather than dichotomous categories, disjunctive (inclusive and exclusive) rather than conjunctive features, and variation not just in N and k, but also in the feature salience parameters. This will require measuring distances on the deeper layers of networks composed of multiple inner layers. With initially complex and *entangled* category representations on the first layer, deeper architectures can learn deeper representations that are *disentangled* and then readily separable by a simple linear projection on a separation plane, as they were in our initial model.

## References

Baldi.P and Hornik. K Neural networks and principal component analysis: Learning from examples without local minima.Neural Networks, 2(1):53–58, 1988.

Bao, S. (2015). Perceptual learning in the developing auditory cortex. European Journal of Neuroscience, 41(5), 718-724.

Bengio, Y., Courville, A, Vincent, P. (2013). Representation learning: A review and new perspectives. *IEEE Trans. PAMI, special issue Learning Deep Architectures*. 35: 1798–1828.

Bishop. C, Pattern Recognition and Machine Learning (Information Science and Statistics),2006. Springer-Verlag New York, Inc. Secaucus, NJ, USA, ISBN:0387310738

*Bourlard, H.; Kamp, Y. (1988). "Auto-association by multilayer perceptrons and singular value decomposition". Biological Cybernetics.* ***59*** *(4–5): 291–294.* doi:10.1007/BF00332918

Cangelosi, A. (2017). Connectionist and Robotics Approaches to Grounding Symbols in Perceptual and Sensorimotor Categories. In Handbook of Categorization in Cognitive Science (Second Edition) (pp. 795-818).

de Leeuw, Joshua, Janet Andrews, Kenneth Livingston, and Benjamin Chin. "The effects of categorization on perceptual judgment are robust across different assessment tasks." Collabra: Psychology 2, no. 1 (2016).

Edwards, D. J. (2017). Unsupervised categorization with a child sample: category cohesion development. European Journal of Developmental Psychology, 14(1), 75-86.

Fisher, R. A. *(1936). "The Use of Multiple Measurements in Taxonomic Problems". Annals of Eugenics.* ***7*** *(2): 179–188*

Folstein, J. R., Palmeri, T. J., Van Gulick, A. E., & Gauthier, I. (2015). Category learning stretches neural representations in visual cortex. Current Directions in Psychological Science, 24(1), 17-23.

Goldstone, R. L., Rogosky, B. J., Pevtzow, R., & Blair, M. (2017). The Construction of Perceptual and Semantic Features During Category Learning. In Handbook of Categorization in Cognitive Science (Second Edition) (pp. 851-882).

Harnad, S. (ed.) (1987) *Categorical Perception: The Groundwork of Cognition*. New York: Cambridge University Press.

Harnad, S. Hanson, S.J. & Lubin, J. (1995) Learned Categorical Perception in Neural Nets: Implications for Symbol Grounding. In: V. Honavar & L. Uhr (eds) Symbol Processors and Connectionist Network Models in Artificial Intelligence and Cognitive Modelling: Steps Toward Principled Integration. Academic Press.

Harnad, S. (2017). To cognize is to categorize: Cognition is categorization. In Handbook of Categorization in Cognitive Science (Second Edition) (pp. 21-54).

*Hopfield, J J (1 April 1982). "Neural networks and physical systems with emergent collective computational abilities". Proceedings of the National Academy of Sciences of the United States of America.* ***79*** *(8): 2554–2558. doi:10.1073/pnas.79.8.2554*

*Hughes, G.F. (January 1968). "On the mean accuracy of statistical pattern recognizers". IEEE Transactions on Information Theory.* ***14*** *(1): 55–63. doi:10.1109/TIT.1968.1054102.*

Hussein, B. A. S. (2012). The Sapir-Whorf Hypothesis Today. *Theory and Practice in Language Studies*, 2(3), 642-646

Jacobs, G. (2013). Comparative Color Vision. Elsevier.

Kay, P., & Kempton, W. (1984). What is the Sapir-Whorf hypothesis? *American Anthropologist*, 86(1), 65-79.

LeCun, Y., Bengio, Y., & Hinton, G. (2015). Deep learning. *Nature*, 521(7553), 436-444.

Pearson, K (1901). On lines and planes of Closest Fit to Systems of Points in Space. Philosophical Magazine. 2 (11) : 559-572

Pérez-Gay Juárez, F., Thériault, C., Gregory, M., Rivas, D. et Harnad, S. (2017). How and why does category learning cause categorical perception? *International Journal of Comparative Psychology*, *30*. Récupéré de https://escholarship.org/uc/item/8rg6c087

Richard O. Duda, Peter E. Hart, David G. Stork, Pattern Classification (2nd edition, 2001), Wiley, New York, 2001 ISBN 0-471-05669-3.

Statistical Pattern Recognition, Second Edition. Andrew R. Webb, John Wiley & Sons, Ltd. 2002, ISBNs: 0-470-84513-9.

*Trunk, G. V. (July 1979). "A Problem of Dimensionality: A Simple Example". IEEE Transactions on Pattern Analysis and Machine Intelligence. PAMI-1 (3): 306–307. doi:10.1109/TPAMI.1979.4766926.*

Vincent,P. Larochelle,H.,Bengio,Y., Menzagol,P-A., Extracting and composing robust features with denoising autencoders.Machine Learning, Proceedings of the Twenty-Fifth International Conference (ICML 2008), Helsinki, Finland, June 5-9, 2008